\pdfoutput=1

\documentclass[sigconf,screen,nonacm]{acmart}

\usepackage{xspace}
\usepackage{latexsym}
\usepackage{amsmath}
\usepackage[T1]{fontenc}
\usepackage{booktabs}
\usepackage{array}
\usepackage[utf8]{inputenc}
\usepackage{microtype}
\usepackage{graphicx}
\usepackage{tabularx}
\usepackage{float}
\newcommand{\sys}{\textsc{GraphRAGProv}\xspace}
\newcommand{\npc}{\textsc{NPCS}\xspace}

\setcopyright{none}
\acmDOI{}
\acmISBN{}

\title{GraphRAGProv: GraphRAG Meets How-Provenance}

\author{Wadie El Amrani}
\affiliation{
  \institution{Independent}
  \city{Toulouse}
  \country{France}
}
\email{elamrani@wadie.fr}

\author{Luis Galárraga}
\affiliation{
  \institution{INRIA}
  \city{Rennes}
  \country{France}
}
\email{luis.galarraga@inria.fr}

\begin{document}

\begin{abstract}
GraphRAG systems often use knowledge graphs to retrieve relevant context for an LLM that produces the final answer. That makes answers more inscrutable and therefore less appealing for domains where transparency is paramount. We present \sys{}, an architecture that executes SPARQL queries over source-grounded RDF graphs to compute answers and their how-provenance polynomials. These polynomials record which statement occurrences were used and how they were combined. Because each occurrence is linked to its source text, users can follow these links to check whether the extracted statements reflect the text and whether the query combines them to answer the question. Querying thereby becomes a way to inspect and evaluate the text-to-graph transformation. Our interactive demo examines 400 cases from two multi-hop QA benchmarks, revealing extraction and query errors even when the final answer is correct.
\end{abstract}
\begin{teaserfigure}
    \centering
    \includegraphics[width=0.9\textwidth]{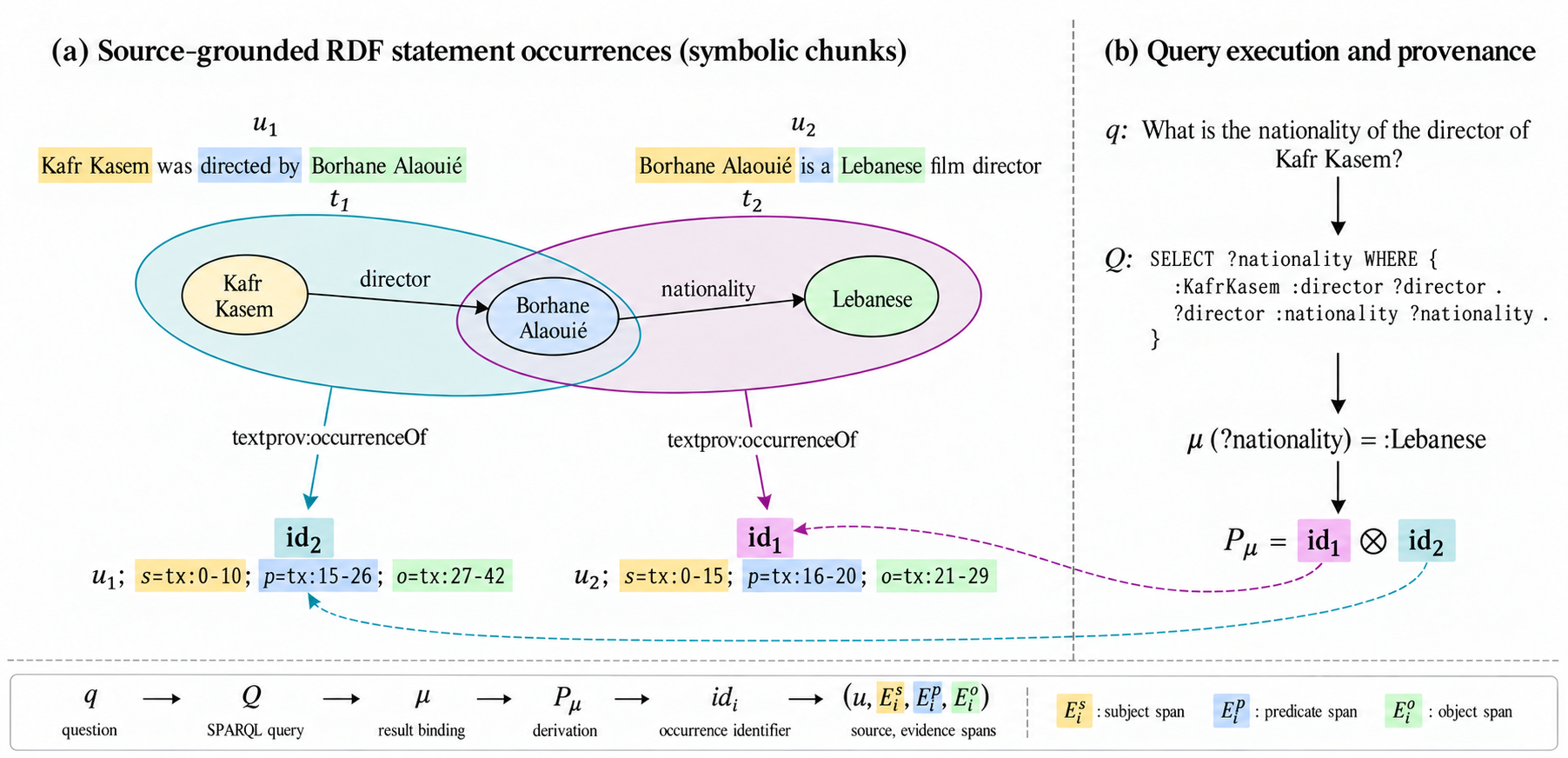}
    \caption{
    (a) Source-grounded RDF statement occurrences constructed from text.
    (b) SPARQL question answering and provenance tracing from answer bindings to RDF occurrences and exact source text evidence.}
    \Description{Diagram linking source-grounded RDF statement occurrences to SPARQL answer bindings, provenance polynomials, and exact supporting text spans.}
    \label{fig:overview}
\end{teaserfigure}
\maketitle

\raggedbottom
\begin{figure*}[t]
    \centering
    \includegraphics[width=0.8\textwidth]{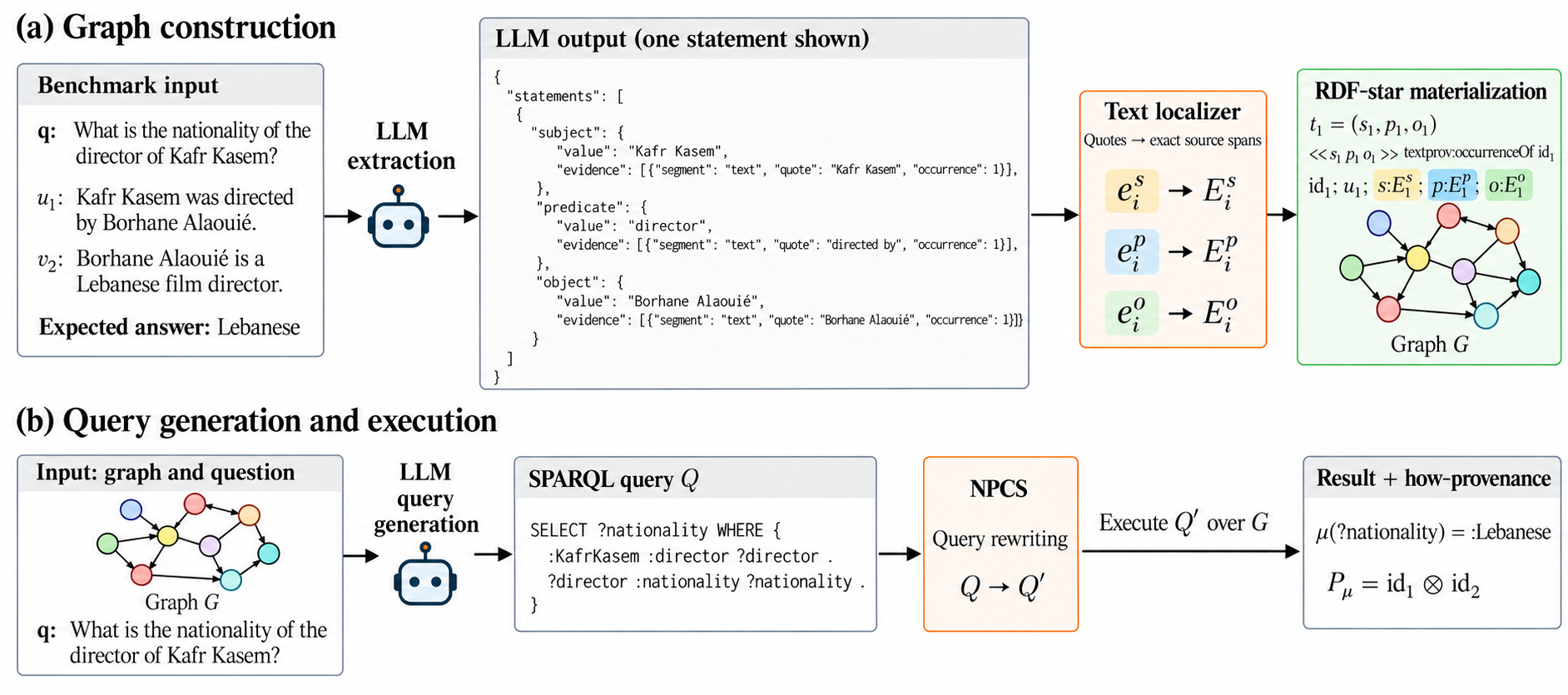}
    \caption{GraphRAGProv architecture.
    An LLM extracts semantic statements and their textual evidence.
    The localizer resolves evidence to exact source spans for RDF-star
    materialization. An LLM generates SPARQL, which NPCS rewrites
    to compute answer bindings and how-provenance.}
    \label{fig:architecture}
\end{figure*}
\section{Introduction}
GraphRAG uses graph structure to connect information across a corpus of text. Microsoft GraphRAG builds entity relation graphs and community summaries, while HippoRAG builds an OpenIE graph and uses Personalized PageRank for retrieval~\cite{edge2024graphrag,gutierrez2024hipporag}. In many such architectures, the graph determines what evidence reaches an LLM, but the LLM still produces the final answer. A retrieved path or subgraph may therefore show what evidence was available to the LLM without specifying how the returned value was computed. This is limiting in critical settings such as medical diagnosis or computational justice, where the computation leading to an answer may itself need to be verified.

One approach is to produce the answer by querying the graph itself. This requires representing facts from the text documents as explicit statements in a data graph, such as subject-predicate-object triples in RDF. A question can then be translated into SPARQL and evaluated by a graph database engine, so that the query result is the final answer~\cite{lan2021kbqa,peng2025graphrag}. While this approach is well established when a knowledge graph already exists~\cite{lan2021kbqa,feng2025cypherbench}, applying it to document collections requires constructing the graph from text first, as in Neo4j's LLM Graph Builder~\cite{neo4j2024llmgraphbuilder} and KAG~\cite{liang2025kag}. KAG reports answering questions from retrieved subgraphs in 33\%, 34\%, and 18\% of cases on HotpotQA, 2WikiMultiHopQA, and MuSiQue, respectively, and supplements graph reasoning with text retrieval when it yields no results. \sys instead uses an LLM to translate questions into SPARQL, with the returned bindings constituting the answers and no text retrieval or LLM generation as a fallback. To make these computations inspectable, we augment query evaluation with database how-provenance ~\cite{green2007provenance}, which identifies the contributing statements and records how they were combined. The database computes both the answers and their derivations, without requiring an LLM to interpret retrieved context. Each returned binding is annotated with a polynomial over statement identifiers: multiplication records joint use, while addition records alternative derivations. \npc~\cite{asma2024npcs} computes these annotations by query rewriting over reified RDF statements.

A derivation identifies the statements used to compute an answer, but does not establish whether those statements are supported by the source text. \sys{} makes this interpretation inspectable by preserving each extraction as a \emph{statement occurrence} linked to its source passage and evidence spans. The extractor supplies quotations for the subject, predicate, and object; a deterministic localizer resolves them to exact character spans, and RDF-star reification preserves the occurrence identifier and its source links. \npc{}~\cite{asma2024npcs} uses these identifiers in provenance polynomials, so following a factor recovers both the assertion used in the computation and the original text against which that assertion can be checked.

Together, query execution, how-provenance, and textual grounding provide a way to inspect and evaluate the text-to-graph transformation. The extractor may misread a passage, entity identifiers may be inconsistent, or the generated SPARQL may fail to capture the question. This explicit chain allows users to examine whether the contributing statements reflect the source text and whether the query combines them as intended.

The present demonstration lets users move across this complete chain : \textbf{question $\rightarrow$ SPARQL $\rightarrow$ answer binding $\rightarrow$ provenance polynomial $\rightarrow$ RDF occurrence $\rightarrow$ exact source spans} and modify either the question or the SPARQL query to observe how the answer and its derivation change. We examine the availability and diagnostic value of this trace on 400 cases from 2WikiMultiHopQA and MuSiQue.
\section{GraphRAGProv}
\label{sec:graphragprov}
\begin{figure*}[t]
\centering
\includegraphics[width=\textwidth]{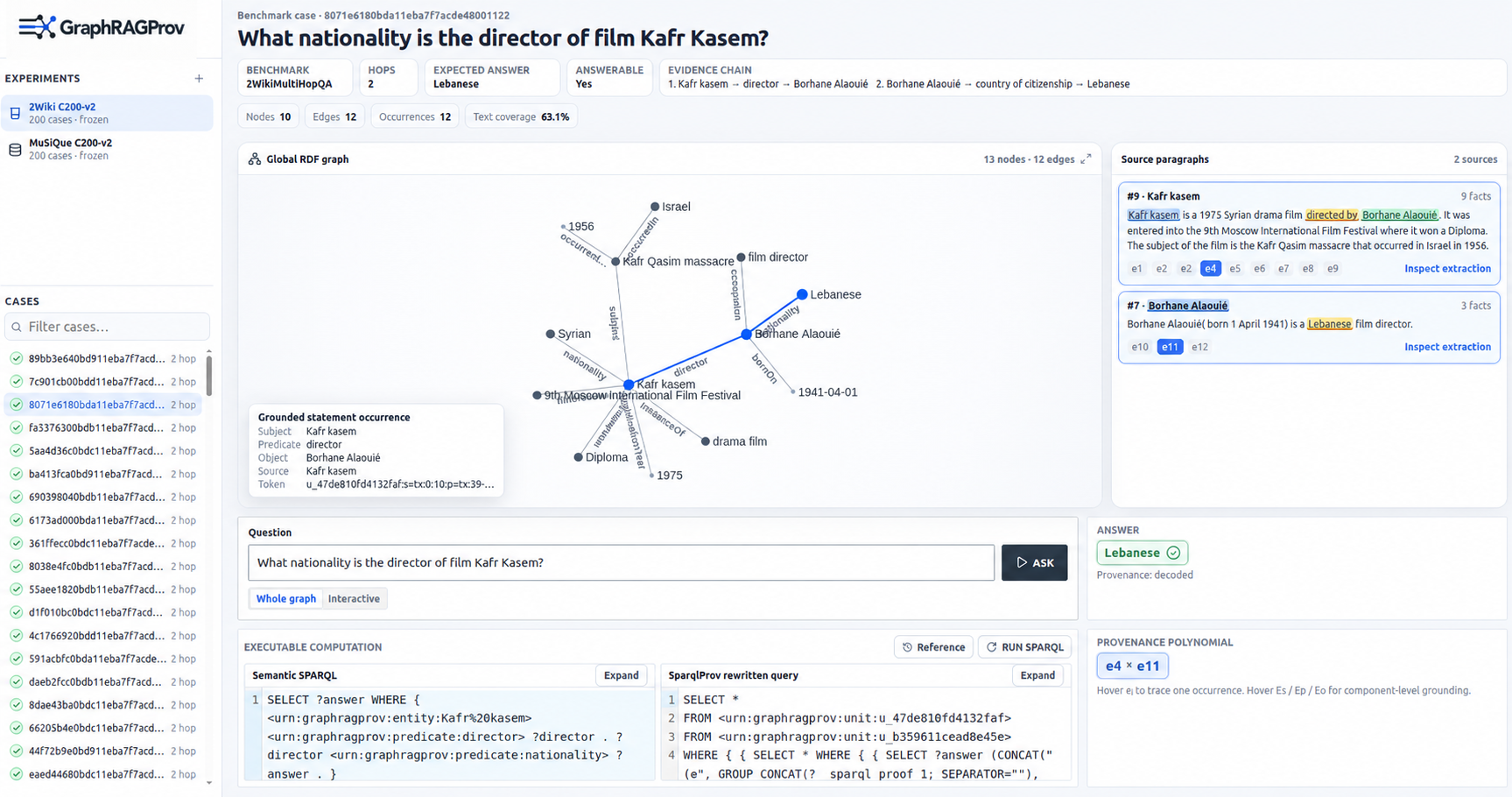}
\caption{\sys interactive workbench. Users can inspect SPARQL queries, follow answer derivations through how-provenance, and highlight source expressions associated with participating RDF statements.}
\label{fig:demo-interface}
\end{figure*}

\noindent\textbf{Graph construction.} Figure~\ref{fig:architecture}(a) illustrates how \sys transforms source paragraphs into an RDF knowledge graph. Extracting relational statements with textual evidence is well established: Stanford OpenIE retains token spans for extracted subjects, relations, and objects~\cite{angeli2015openie}, while recent LLM-based approaches such as AEVS associate semantic triples with character-level evidence~\cite{yang2026aevs}. Following this line of work, an LLM interprets natural-language expressions as RDF statements $t_i=(s_i,p_i,o_i)$. For example, ``Borhane Alaouié is a Lebanese film director'' can yield $(\text{Borhane Alaouié},\text{nationality},\text{Lebanese})$. The relation \texttt{nationality} is inferred from the adjective ``Lebanese'' and ''is a'', rather than copied from the sentence. Alongside each statement, the LLM provides textual quotations supporting its components. A deterministic localizer resolves these quotations to exact source spans $E_i^s,E_i^p,E_i^o$. Quotations that cannot be located trigger a correction request to the LLM; unresolved quotations cannot be accepted as exact evidence.

\noindent\textbf{Symbolic chunk.} We call a source-grounded statement occurrence a \emph{symbolic chunk}. Each occurrence receives an identifier $id_i$, connecting its RDF triple to the source paragraph $u_i$ and the available evidence spans. Identical triples extracted from different passages retain separate occurrence identifiers and textual evidence. GraphRAGProv materializes these connections using reification with RDF-star through \texttt{textprov:occurrenceOf}, as illustrated in Figure~\ref{fig:overview}(a).

\noindent\textbf{Query execution.} Figure~\ref{fig:architecture}(b) shows the answering process. Given a question $q$ and the constructed graph, an LLM generates a SPARQL query $Q$, which GraphDB executes to obtain answer bindings $\mu$. NPCS~\cite{asma2024npcs} rewrites the query to additionally compute how provenance $P_\mu$. For the Kafr Kasem example, the database returns \emph{Lebanese}, with $P_\mu=id_1\otimes id_2$. Multiplication represents the joint use of statement occurrences, while addition represents alternative derivations~\cite{green2007provenance}. Each provenance identifier resolves to its source evidence, completing the trace in Figure~\ref{fig:overview}(b):
\[
q \rightarrow Q \rightarrow \mu \rightarrow P_\mu \rightarrow id_i \rightarrow (u_i,E_i^s,E_i^p,E_i^o).
\]
\section{Demonstration}
\label{sec:demo}
The \href{https://graphragprov.wadie.fr}{live demonstration} is publicly accessible, and the \href{https://github.com/wadie999/GraphRAGProv}{source code} is available under the MIT license.

\noindent\textbf{User Experience.}
Figure~\ref{fig:demo-interface} shows the GraphRAGProv workbench. Users can explore recorded cases from 2WikiMultiHopQA and MuSiQue, each containing the benchmark question, source paragraphs, and constructed RDF graph. Selecting a graph statement highlights its supporting text and vice versa. Users can inspect the LLM-generated SPARQL query, execute it, and examine the returned answers and how-provenance polynomials. Selecting a provenance factor reveals the contributing statement and its textual evidence.
The workbench also allows editing natural-language questions or SPARQL queries to observe changes in answers and derivations.

\noindent\textbf{Implementation.}
GraphDB stores the RDF graphs, with named graphs separating source paragraphs. The React frontend communicates with a FastAPI backend. SPARQL queries are rewritten using a Rust implementation of NPCS\footnote{\url{https://github.com/wadie999/sparql-prov}} to compute answer bindings and how-provenance. Graph construction and query generation use DeepSeek-V4-Flash. The extraction\footnote{\url{https://github.com/wadie999/graphragprov/blob/main/src/graphragprov/extraction/prompt.py}} and query-generation\footnote{\url{https://github.com/wadie999/graphragprov/blob/main/src/graphragprov/qaagent/prompt.py}} prompts can be found in the source code.
\section{Evaluation}
\label{sec:evaluation}
\noindent\textbf{Experimental setting.} We examine 200 compositional questions from 2WikiMultiHopQA and 200 answerable questions from Musique. For each case, the graph is constructed from its designated supporting passages, excluding document retrieval from the experiment. Graph construction and SPARQL generation use DeepSeek-V4-Flash, with access to the benchmark question, expected answer, and reasoning evidence. This supervised setting allows us to examine the resulting graphs and computations, but does not establish generalization to unseen questions. We retain the first query that passes mechanical validation and executes, without retrying based on answer correctness.

\noindent\textbf{Trace availability.} Table~\ref{tab:coverage} reports the number of cases reaching each stage. A complete trace requires every factor in the decoded derivations to resolve to its statement occurrence and matching source spans. Complete traces are available for 193 of 200 2Wiki cases and 118 of 200 MuSiQue cases. All 311 decoded traces pass the occurrence and source-span checks, and their answer bindings match ordinary query execution. Sixteen MuSiQue cases return an answer without decoded provenance.

\noindent\textbf{Case inspection.} Table~\ref{tab:audit-cases} illustrates what can be learned by inspecting the computation and its sources. In Meesaya Murukku, the answer is correct and the cited word ``Indian'' exists at the recorded offsets, but it describes the film rather than its director. The error lies in the interpretation of the text, not in provenance resolution. Labyrinth reveals a different problem: the query returns the correct year using two individually supported assertions, yet never establishes that Acornsoft published the game. Its provenance faithfully records the executed computation, but that computation does not express the question. In contrast, the Caledon River case yields a sufficient derivation from one passage containing both required relations, despite the benchmark designating two supporting passages.
\begin{table}
\centering
\small
\setlength{\tabcolsep}{4pt}
\begin{tabular}{lrrr}
\toprule
\textbf{Stage} & \textbf{2Wiki} & \textbf{MuSiQue} & \textbf{Total} \\
\midrule
Cases & 200 & 200 & 400 \\
SPARQL executed & 200 & 198 & 398 \\
Answer returned & 193 & 134 & 327 \\
Provenance decoded & 193 & 118 & 311 \\
Complete text trace & 193 & 118 & 311 \\
\bottomrule
\end{tabular}
\caption{Answer-to-source trace availability.}
\label{tab:coverage}
\end{table}

\noindent\textbf{Missing answers.} An empty result provides no answer derivation to inspect. In Aldobrandino, the father relation and birth date are present in the graph, but attached to different IRIs representing the same person. The resulting join fails despite the required information being extracted. This failure is identified by inspecting the graph and query directly, rather than following provenance of a nonexistent result.
\begin{table}[H]
\centering
\footnotesize
\setlength{\tabcolsep}{3pt}
\renewcommand{\arraystretch}{1.05}
\begin{tabularx}{\columnwidth}{
    @{}
    >{\raggedright\arraybackslash}p{0.38\columnwidth}
    >{\raggedright\arraybackslash}X
    @{}
}
\toprule
\textbf{Question and case ID} & \textbf{Inspection finding} \\
\midrule

\textbf{Meesaya Murukku} (2Wiki)\newline
Which country is its director from?\newline
{\tiny\texttt{\detokenize{ba413fca0bd911eba7f7acde48001122}}}
&
Correct answer, \emph{India}, but the cited text describes the film as Indian, not its director.
\par\smallskip
{\scriptsize\(\begin{aligned}[t]
&q\to Q\to\mu\to P_\mu\to\textcolor{red}{id_i}\\[-1pt]
&\quad
\end{aligned}\)}
\\
\addlinespace[4pt]

\textbf{Labyrinth} (MuSiQue)\newline
When did its publisher cease operating?\newline
{\tiny\texttt{\detokenize{2hop__170823_120171}}}
&
Correct answer, 1986, but the query omits the game--publisher relation.
\par\smallskip
{\scriptsize\(\begin{aligned}[t]
&q\to\textcolor{red}{Q}\to\mu\to P_\mu\to id_i\\[-1pt]
&\quad
\end{aligned}\)}
\\
\addlinespace[4pt]

\textbf{Aldobrandino} (2Wiki)\newline
When was his father born?\newline
{\tiny\texttt{\detokenize{96a72db40bdd11eba7f7acde48001122}}}
&
Two IRIs represent the same father, preventing the join.
\par\smallskip
{\scriptsize\(\begin{aligned}[t]
&q\to Q\to\textcolor{red}{\mu=\varnothing}\to\textcolor{gray}{P_\mu\to id_i}\\[-1pt]
&\quad\textcolor{gray}{}
\end{aligned}\)}
\\
\addlinespace[4pt]

\textbf{Caledon River} (MuSiQue)\newline
Where is the source of the receiving river?\newline
{\tiny\texttt{\detokenize{2hop__207597_63853}}}
&
One passage supports both required relations, although the benchmark designates two.
\par\smallskip
{\scriptsize\(\begin{aligned}[t]
&q\to Q\to\mu\to P_\mu\to id_i\\[-1pt]
\end{aligned}\)}
\\

\bottomrule
\end{tabularx}
\caption{Representative cases. Red identifies an unsupported assertion, an incorrect query, or an empty result; gray indicates unavailable provenance stages. Questions are abbreviated, and case IDs identify the saved experiments.}
\label{tab:audit-cases}
\end{table}

\section{Conclusion}
\label{sec:conclusion}
An executable query answers what it asks of a graph, not necessarily what the user asked of the original text: answering the original question also requires the graph to preserve the relevant knowledge and the query to express the intended computation. \sys makes this distinction inspectable. SPARQL specifies the computation, how-provenance identifies the contributing statement occurrences, and symbolic chunks connect them to their textual origins. These connections allow users to inspect the interpretations made by the LLM during graph construction. Our demonstrations show why these layers must be examined together: supported statements may participate in an incorrect query, correct answers may depend on unsupported interpretations, and missing answers may result from inconsistent entity representations rather than absent knowledge. Provenance exposes the chain to help users locate and diagnose failures.

\bibliography{anthology,custom}
\bibliographystyle{ACM-Reference-Format}

\appendix
\end{document}